\documentclass[11pt]{article}
\usepackage[T1]{fontenc}
\usepackage{mathptmx}
\usepackage[margin=1in]{geometry}
\usepackage{graphicx}
\usepackage{amsmath,amssymb}
\usepackage{booktabs}
\usepackage{xcolor}
\usepackage{microtype}
\usepackage[numbers,sort&compress]{natbib}
\usepackage[hyperfootnotes=false]{hyperref}
\definecolor{linkblue}{rgb}{0.21,0.49,0.74}
\hypersetup{colorlinks,allcolors=linkblue,
  pdftitle={CADMAGEDDON: High-Fidelity CAD Reconstruction with Local Inference},
  pdfauthor={Oleg Pavlov, Nikita Gavrilov, Soslan Kabisov, Gennadii Savrasov, Andrey Kuznetsov, Anton Konushin, Dmitrii Zhemchuzhnikov}}

\title{CADMAGEDDON: High-Fidelity CAD Reconstruction with Local Inference}
\author{
Oleg Pavlov\textsuperscript{1,4}\quad
Nikita Gavrilov\textsuperscript{1}\quad
Soslan Kabisov\textsuperscript{1}\quad
Gennadii Savrasov\textsuperscript{1,4}\\
Andrey Kuznetsov\textsuperscript{2,3}\quad
Anton Konushin\textsuperscript{1,4}\quad
Dmitrii Zhemchuzhnikov\textsuperscript{1,4,\dag}\\[0.6em]
{\normalsize \textsuperscript{1}Lomonosov Moscow State University\qquad
\textsuperscript{2}Innopolis University\qquad
\textsuperscript{3}FusionBrain Lab}\\
{\normalsize \textsuperscript{4}AXXX, Moscow, Russia}\\[0.3em]
{\small \textsuperscript{\dag}Corresponding author: \texttt{zhemchuzhnikovds@my.msu.ru}}
}
\date{}

\newcommand{\tableonehook}{\scriptsize\setlength{\tabcolsep}{0.9pt}}

\begin{document}
\maketitle

\begin{abstract}
CAD practitioners often work with proprietary parts that they are unwilling to share with external services. For individuals and small companies, bringing this work in-house also requires computational resources they can afford. We introduce CADMAGEDDON, an open-source agent that reconstructs editable CAD from meshes and can run locally on single-GPU systems and current-generation laptops. The method decomposes this complex task into a chain of simple visual decisions about what to construct and what to correct. Geometric tools support precise construction, allowing the agent to focus on visual reasoning. Using the mid-sized Qwen3.8-27B model, CADMAGEDDON outperforms prior approaches, including systems based on frontier models, in reconstruction quality on all the main benchmarks. CADMAGEDDON integrates with FreeCAD, bringing its reconstructions into practitioners' existing CAD workflows. Code: \url{https://github.com/kulibinai/cadmageddon}.
\end{abstract}

\section{Introduction}
\label{sec:intro}

Much of the geometry that engineers need to work with is available only as a mesh: an export from a legacy system, or a file stripped of its feature tree on the way from a supplier. Meshes represent a part as a finite set of facets and lack the feature-level structure and design intent that support parametric redesign; their geometric fidelity is limited by the tessellation. Reconstructing a part from its mesh as an executable, feature-based CAD model can infer a plausible sequence of operations --- sketches, extrusions, revolutions and holes --- that a CAD system can rebuild and an engineer can readily edit. Today, this process is mostly done by hand.
There is a need for an automatic replacement with two capabilities: accurately reconstructing real mechanical parts, including geometries more complex than the sketch--extrude shapes that dominate common training data; and running on premises on a single machine, without requiring a cluster or an external API. Local execution matters because these parts are often proprietary, and their owners may be unwilling to upload them to an outside service.

Single-pass learned models that translate point clouds or images into CAD code~\citep{cadrecode,cadrille,cadevolve,cadcodervlm} run locally, but where they have been evaluated on real mechanical parts, they lose much of their accuracy~\citep{cadena}, and iterative refinement~\citep{cadreasoner} does not prevent the drop. Stepwise generation~\citep{cadena} reduces this loss without eliminating it. These learned models are trained on programs whose numerical values are quantised or rounded, and their precision is further limited by how finely the input point cloud or image is encoded and by how well a language model can predict numerical values. Direct geometric fitting~\citep{cadfit}, by contrast, recovers parameter values as precisely as the mesh allows, but its decomposition of the part into operations can misread the design intent, and it leaves roughly a quarter of the parts without a valid reconstruction~\citep{cadena}. Agentic systems such as CADFather~\citep{cadfather} combine a learned generator, geometric fitting and parameter optimisation (Sec.~\ref{sec:related}). Frontier LLMs that write and execute code in a sandbox report strong results on BenchCAD~\citep{benchcad,benchcadboard}, but these figures are self-reported, conditioned on images rather than meshes, and computed on subsets of at most 1000 parts or of unstated size, and the models run only as a remote service.

The limited numerical precision of learned models and the misread design intent of direct fitting suggest separating the two halves of each feature: its numerical parameters should be fitted to the target rather than predicted, and its structure should be decided visually. Interactive scan-to-CAD workflows already separate them, with an engineer choosing each feature and fitting supplying its values.

CADMAGEDDON makes the same split for every feature, with a vision--language model in place of the engineer (Fig.~\ref{fig:loop}). Geometric algorithms analyse the target mesh and name the entities the model needs to refer to --- cutting planes, levels along a plane normal, fitted lines and arcs, repeated hole patterns --- and render the target and the current reconstruction in a shared colour convention. The model composes each feature from these names, choosing its plane, level, profile primitives and operation. The tools resolve the names to fitted parameters, emit and execute the corresponding CAD program, and report which volume is still missing or in excess, so that the next decision either adds a feature or corrects one. The model refers to geometry primarily by name, and the values of the features it builds come from fitting; where no detector has named a location, the model can compose one from existing names.

This design needs neither frontier scale nor a trained reconstruction model. CADMAGEDDON uses a single open-weight model, Qwen3.8-27B~\citep{qwen38}. With it, CADMAGEDDON reaches the highest IoU and GMS and the lowest Chamfer distance on DeepCAD, Fusion~360, MCB, BenchCAD and CADENA-Bench (Table~\ref{tab:main}); GMS~\citep{cadena} scores how well the reconstructed surfaces match the target's surface types, which volume-based IoU does not capture. It is ahead of CADFather~\citep{cadfather}, which runs the same model, and of prior learned and optimisation-based methods. The gains are largest where surface structure matters. On the mechanical parts of MCB and CADENA-Bench, GMS rises from CADFather's 0.705 and 0.800 to 0.818 and 0.909 under the same evaluation. On BenchCAD it also scores above the self-reported entries of frontier models on the public leaderboard (Table~\ref{tab:leaderboard}), although inputs and subsets differ. With FP8 weights, the model fits on a single 48\,GB GPU. The same agent can also write native FreeCAD programs, in addition to CadQuery (Sec.~\ref{sec:exp:benchcad2}), so its reconstructions enter FreeCAD workflows directly.

Our contribution is CADMAGEDDON, an open-source mesh-to-CAD agent that turns reconstruction into a chain of visual decisions over named geometric entities. Algorithmic tools detect these entities on the target mesh and supply their fitted parameters, while a single mid-sized open-weight model composes each feature from them and decides what to correct, with no trained reconstruction model. The agent runs on a single GPU, writes both CadQuery and native FreeCAD programs, and reconstructs more accurately than prior learned and optimisation-based methods on five benchmarks.

\section{Related Work}
\label{sec:related}

\paragraph{Generative and structured CAD models.}
One line of learning-based CAD modelling generates construction sequences. DeepCAD models sketch--extrude sequences, and its dataset is the corpus and benchmark of much later work~\citep{deepcad}; later models factorise such sequences or generate their constrained sketches~\citep{skexgen,sketchgen,vitruvion,cadvlm}, built mainly to synthesise designs, not to reconstruct a given part; recent text-to-CAD work also supervises the design intent that keeps programs editable~\citep{intent2cad}. Other work recovers structured geometry via boundary representations~\citep{complexgen,solidgen,brepgen}, CSG trees of primitives~\citep{csgnet,ucsgnet,caprinet}, and extrusion cylinders or sketch--extrude bodies combined by Booleans~\citep{point2cyl,extrudenet,secadnet}; none of these yields general feature programs.

\paragraph{Shape-to-program reconstruction.}
Learned models now map an observed shape directly to code. CAD-Recode translates point clouds into CadQuery with a fine-tuned language model~\citep{cadrecode}, cadrille adds image and text input and reinforcement learning on geometric rewards~\citep{cadrille}, CAD-Coder writes CadQuery from images~\citep{cadcodervlm}, and CADEvolve, PLLM and GIFT enlarge or self-generate the training data~\citep{cadevolve,pllm,gift}. Newer variants add primitive-aware tokenisation for point clouds~\citep{primitivecad}, pretrained multi-view geometric priors~\citep{vggtcad}, a construction-plan interface shared across input modalities~\citep{miragecad} and multi-task training across modalities~\citep{unicad}. Each emits the full program, including all parameter values, in a single pass. Others move towards real inputs: CADonatello segments multi-part noisy scans into parts and derives extrusion parameters from the recovered sections~\citep{cadonatello}, and SightCAD handles occluded views~\citep{sightcad}. Stepwise methods instead execute the program after each construction step. The Fusion~360 Gallery pairs human-authored sketch--extrude histories with a step-by-step execution environment and recovers target programs by neurally guided search~\citep{fusion360}. Zone-graph methods search over sketch, extrude and Boolean sequences, scoring candidate extrusions against the target with a graph neural network~\citep{zonegraphs}. CADFit fits and validates operations incrementally by hybrid optimisation under an IoU objective, pruning candidates with a learned sketch prior~\citep{cadfit}. StepCAD combines a state-conditioned language-model policy, which proposes construction actions from the target and the intermediate geometry, with an IoU-guided tree search over local edits~\citep{stepcad}. CADReasoner regenerates the whole program over several rounds conditioned on its discrepancy with the input~\citep{cadreasoner}, CADENA appends one operation at a time while comparing renders of the target and the current build~\citep{cadena}, and SOV-CAD reconstructs stepwise from orthographic views with an offline-RL Decision Transformer~\citep{sovcad}. Differentiable CAD programs and CSG rasterisation refine the parameters of a fixed structure after the fact~\citep{differentiablecad,diffcsg}. In most learned methods a model writes the values; in the search-based ones a search against the target selects the structure, in CADFit together with the values. In CADMAGEDDON a vision--language model chooses only the structure of each feature, and fitting supplies its values.

\paragraph{Tool-using CAD agents.}
Many recent agents give language and vision--language models tools inside an executable CAD environment. CAD-Assistant, a training-free framework, lets a vision--language model plan over CAD-specific tools that include rendering and a cross-section generator, executes them in FreeCAD, and adapts later actions to their effect on the geometry~\citep{cadassistant}. IterCAD trains a multimodal agent with progressive supervised fine-tuning and geometry-aware reinforcement learning to generate and edit CAD code over rounds of execution~\citep{itercad}, and ToolCAD and COSMO-Agent train language models as tool users over CAD kernels and simulation engines~\citep{toolcad,cosmoagent}. Further agents generate parts from text or images and add kernel-measured validation, trace-guided repair, editable intermediate representations or distilled expert procedures~\citep{cadsmith,tracecad,cadir,artisancad}. Text-to-CAD agents increasingly reason about executed geometry, learning to critique the built state~\citep{racad}, anticipating the geometry an operation should produce~\citep{cadwam}, planning over a graph of operations~\citep{goprcad} or correcting each other in multi-agent loops~\citep{craft}, and candidates can be selected by consensus instead of a verifier~\citep{cadtts}. CADForge reconstructs CadQuery from a single image by decomposing the object into components and reasoning about each one's geometry~\citep{cadforge}. Embodied CAD lets agents act through a stratified library of kernel-checked CAD skills~\citep{embodiedcad}, open skill libraries wrap engineering tools for agents~\citep{meskills}, and agents' reports of success are poorly calibrated against kernel-checked requirements~\citep{doneisnotcorrect}. Beyond mechanical parts, agents reconstruct as-built building models from reality-capture data~\citep{agent2bim}, and parametric building components are recovered from point clouds~\citep{crab}. DeciCAD argues that the action space a harness exposes shapes both the behaviour and the learned policies of CAD agents~\citep{decicad}. What distinguishes our system is therefore not tool use with execution feedback, nor a semantic action space as such, but an action space grounded in the target: our model acts primarily by choosing named entities fitted on the target mesh, and the values of the features it builds come from that fitting. The closest prior system, CADFather~\citep{cadfather}, runs the same frozen Qwen3.8-27B model over a pool of candidate programs, inspecting renders and execution feedback and calling three tools: a learned operation generator, algorithmic proposals fitted to cross-sections of the target, and numerical parameter optimisation. Its model selects candidates and tools and accepts or rejects whole proposed operations. Its authors report a mean GMS well below the mean IoU on MCB and CADENA-Bench (0.766 and 0.721), which they read as volume being reproduced without the surface detail, although the IoU is averaged only over targets where it is defined; and without its fitted proposals, 12\% of a 1000-part MCB subset are left invalid, mostly because no sampled first operation forms a closed solid~\citep{cadfather}. CADMAGEDDON has no generator, optimiser or pool: a single model composes each feature itself, along one trajectory in which features can be undone or dropped.

\paragraph{Fitting and referring to geometry.}
Classical reverse engineering segments a scan, fits planes, cylinders, cones, spheres and tori to its regions, often with RANSAC-style estimators~\citep{efficientransac}, and recognises features such as holes and patterns; in scan-to-CAD software an engineer then combines these fits into a feature-based model. Our instruments belong to this tradition, and the fitting itself is not our contribution. CADFit and the algorithmic proposals of CADFather already turn fits into operations, by optimisation-based search and by fixed geometric procedures respectively~\citep{cadfit,cadfather}; in CADMAGEDDON a vision--language model makes these decisions. Referring to geometry instead of writing it also has precedents. Pointer-CAD builds CAD models from text step by step and selects the edges and faces an operation acts on through pointers into the current B-Rep~\citep{pointercad}. Pointer-CAD~v2 first produces a design plan whose metric parameters are kept in a dictionary, and points to these entries during construction~\citep{pointercadv2}. DIP-CAD, an agent harness for parametric modelling from images, binds commands to faces, edges and vertices through explicit identifiers~\citep{dipcad}. In all three, references point into the model under construction or into a plan, and their values come from the prompt or the model's prediction, not from measurement of a target. Our names point into the target mesh (cross-section levels, fitted lines and arcs, hole stacks, patterns) and resolve to fitted values. In visual prompting, marks on candidate image regions likewise let a vision--language model choose a label instead of producing coordinates~\citep{setofmark}. Our elevation renders mark named levels in this way, but the alternatives are fitted entities rather than image regions.

\paragraph{Benchmarks.}
The test splits of DeepCAD and Fusion~360~\citep{deepcad,fusion360} are the common standard for reconstruction, but both consist of sketch--extrude construction sequences. Benchmarks of real mechanical parts are not yet established: MCB~\citep{mcb}, a corpus collected for classification and retrieval, and CADENA-Bench, which draws deduplicated parts from MCB and two other corpora, have so far been used for reconstruction mainly by CADENA and CADFather~\citep{cadena,cadfather}. BenchCAD evaluates frontier models on industrial part families and studies fine-tuning~\citep{benchcad}, and CADBench compares CAD-specialised and general-purpose systems on meshes and images~\citep{cadbench}. Agentic benchmarks evaluate CAD authoring in FreeCAD~\citep{pcadbench3,cadworld}, and industrial ones pair real design intents or toleranced drawings with programs~\citep{realcadbench,omnimech}.

\section{Method}
\label{sec:method}

\begin{figure*}[t]
\centering
\includegraphics[width=\textwidth]{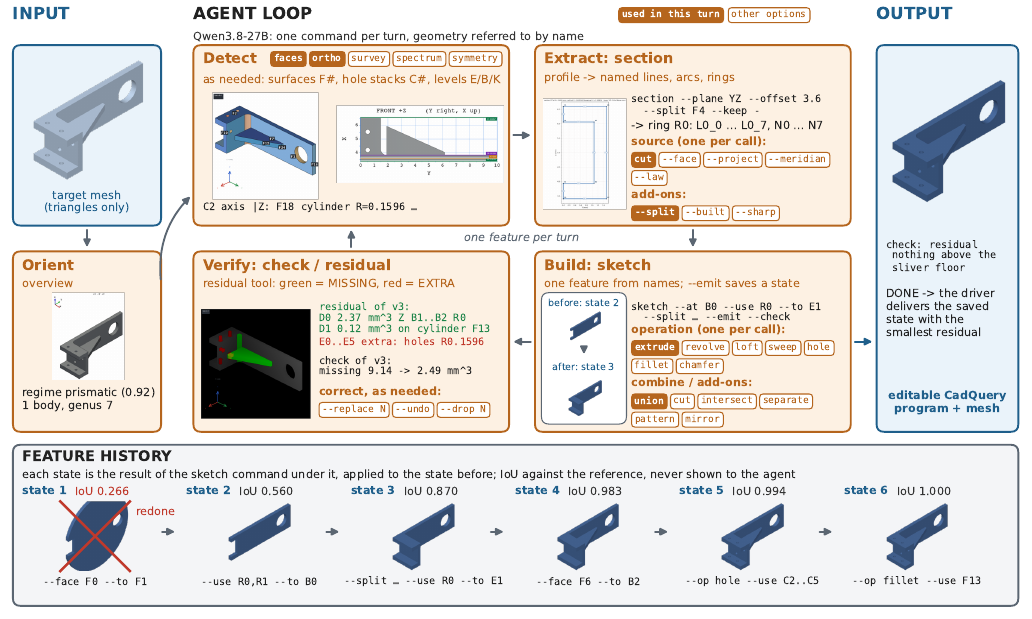}
\caption{The CADMAGEDDON agent loop, shown on one real turn of a Fusion~360 run (state~2 to state~3). Qwen3.8-27B issues one command per turn and refers to geometry mainly by name. It orients itself, detects named entities as needed, extracts a profile, builds one feature, and verifies it against the residual (green: missing, red: extra), correcting features when needed. The driver returns the saved state with the smallest residual. Highlighted options are those used in this turn. The residual render was made afterwards for illustration (the agent read it as text), and \texttt{--split F4} replaces the equivalent value $Y{=}1.4894$ that the agent copied from the \texttt{faces} output. Bottom: the part's feature history. State~1 was redone. The IoU is never shown to the agent.}
\label{fig:loop}
\end{figure*}

CADMAGEDDON takes a triangle mesh of a single part and returns an executable CadQuery program whose solid reproduces it. The reconstruction loop combines geometric analysis, model-guided decisions, and CAD construction. Geometric \emph{instruments} analyse the mesh and the current reconstruction. A vision--language model reads their reports and chooses what to do next. A program builder turns each of the model's choices into CAD operations and executes them. The instruments detect the quantities the model needs and give them short names. The model primarily refers to these named geometric entities fitted to the target mesh; some arguments, such as section offsets, also accept explicit numerical values. The model therefore decides the \emph{structure} of the reconstruction --- which planes, which profiles, which operations, in what order --- while its \emph{parameters} come from numerical fitting to the target. Fig.~\ref{fig:loop} gives an overview of the loop.

\subsection{Preprocessing}
\label{sec:method:preprocess}
Before reconstruction, we repair mesh defects, align the dominant planar face normals with the coordinate axes, and rescale very small parts. The alignment defines the coordinate planes used by all later sections. The emitted program records the rotation and scale so that exported geometry can be mapped back to the input frame.

\subsection{Named geometric entities}
\label{sec:method:names}
The instruments expose the geometry of the target through a small naming grammar. A name consists of a prefix letter for the entity kind, an index, and optional sub-index and qualifier. Named entities include:
levels along a plane normal at which the cross-section changes, such as part ends, boundaries, taper ends and turning points (\texttt{E}, \texttt{B}, \texttt{K}, \texttt{T});
plateaus of constant cross-section (\texttt{c});
closed rings of a fitted cross-section and their vertices, lines and arcs (\texttt{R}, \texttt{N}, \texttt{L}, \texttt{A});
analytic surfaces of the mesh --- planes, cylinders, cones, spheres, tori and surfaces of revolution (\texttt{F});
coaxial hole stacks (\texttt{C}), helices (\texttt{H}) and sweep paths (\texttt{W});
detected patterns and arrays (\texttt{P}) and symmetry candidates (\texttt{S});
and the missing and extra bodies of the current residual (\texttt{D}, \texttt{E}) and the corresponding regions on a cross-section (\texttt{M}, \texttt{X}). The letter \texttt{E} thus denotes both a part-end level and an extra body; the command in which it appears makes the meaning unambiguous.
In \texttt{sketch --use}, a line or arc name selects its enclosing ring. Names are resolved in the context of the current section. When the model needs a location that no detector has named, it can compose one from existing names, for example the midpoint of two vertices or of two levels.

The values behind the names are obtained by fitting. Cross-section polylines are split at corners and recursively decomposed into line and arc segments, using a fitting tolerance estimated from the mesh tessellation. Equal circles are grouped into bolt-circle patterns and linear arrays. When adjacent slices of a plane sweep bracket a step change in cross-section, we refine its location by bisection. The initial slice spacing still limits which changes can be detected. Analytic surfaces are grown from planar patches and fitted to the mesh vertices, again with a tolerance derived from the mesh.

\subsection{Instruments}
\label{sec:method:instruments}
The model drives a command-line interface; it selects inspection tools as needed and can revisit them as the reconstruction evolves. \texttt{overview} renders four isometric and six orthographic views of the target and reports its size, volume, bodies, helices, revolve axis, and axial spans with nearly constant cross-sections. It also reports a heuristic construction regime (helical, revolved, prismatic, patterned or freeform) based on helicity, rotational symmetry, and the prevalence of planar faces; the regime is evidence for the model and does not restrict its choices. \texttt{survey} scores the three principal planes by how prismatic the part is along each normal. \texttt{spectrum} reports intervals of nearly constant cross-section and transitions along the selected plane normal. \texttt{ortho} marks named section levels on four orthographic side views. \texttt{section} extracts a profile and decomposes it into named rings, lines and arcs. The profile is usually a planar cut at a named level, but it can also be the outline of a named face, a projection, a meridian of a revolved part or a fitted law. The command also summarizes the intervals between consecutive levels, reporting cross-section area and loop and hole counts. Additional instruments list analytic faces and hole stacks (\texttt{faces}), detect mirror planes and rotational symmetries (\texttt{symmetry}), and split the difference between target and reconstruction into named missing and extra bodies (\texttt{residual}). Each instrument prints a short text report. Where a picture helps the decision, it also renders images, which are passed to the model.

\subsection{Construction and correction}
\label{sec:method:construct}
A single command, \texttt{sketch}, adds one feature. The model names a plane, a level, the primitives that form the profile and an operation, and optionally the level at which the operation ends. Supported operations are extrusion (optionally tapered or symmetric), revolution, lofts between several levels (including twisted lofts), sweeps along a helix or a detected path, holes along a detected stack, and blends. The builder resolves the names to fitted values and emits the corresponding CadQuery code. It combines the new feature with the current solid by union, subtraction or intersection, or keeps it as a separate body. The model can also request a working frame aligned with an inclined face or axis; the builder maps the resulting operations back to the part frame. The builder executes the candidate program, checks the resulting geometry for validity, and measures how much material is missing or in excess (the \emph{residual}). When the volume discrepancy is small, it also checks for local surface deviations that may have little effect on volume. Residual components smaller than 0.01\% of the target volume are excluded from the missing and excess volume totals; a \emph{clean} residual also requires that these fragments total at most 0.2\% of the target volume. When either mesh is not closed, an occupancy-based measure replaces the Boolean one. The model sees the current state visually by requesting \texttt{progress}, which renders the target and the reconstruction from the same views coloured by their distance to each other, or by sectioning the reconstruction against the target. Features can be dropped, undone, mirrored or patterned, so the next decision may either construct a new feature or correct an existing one.

\subsection{Agent loop}
\label{sec:method:loop}
Fig.~\ref{fig:loop} shows the loop end to end. The system prompt includes a reconstruction guide and a reference to the available commands and options. The model initially receives no views of the part. The driver executes at most one command per turn and returns its text output and any renders to the model. To keep the conversation within the context of a mid-sized model, earlier reasoning is truncated, older command outputs are replaced by their summaries, and only a bounded number of recent renders is kept. The model can reprint an earlier output in full with \texttt{recall}, which it does on 21\% of parts. Turns without a command do not count against the call budget. A reply truncated by the token limit is retried with a larger limit rather than counted as a failure.

When the model declares \texttt{DONE}, the driver asks it to continue once per saved version if mismatched patches cover at least 1\% of the target surface. Once this rule is satisfied, a clean residual (Sec.~\ref{sec:method:construct}) qualifies as complete. With a nonzero residual, the driver may stop with an incomplete reconstruction if the residual has stopped improving or is at most 1\% of the target volume. The run also ends when the command budget is spent or when the driver exhausts its retry budget for missing, failed, or repeated commands. The builder saves a program version when requested. The driver selects among eligible saved versions by minimizing the sum of missing and excess volume relative to the input mesh; no ground-truth model is used.

\subsection{Implementation}
\label{sec:method:impl}
The agent uses Qwen3.8-27B with FP8 weights, served with vLLM through an OpenAI-compatible API and sampled at temperature $0.2$. Each part gets a single attempt, except on BenchCAD-2, where we report the best of up to five completed attempts per part, selected by the score. The command budget depends on the benchmark: 30 commands for DeepCAD and BenchCAD, whose parts are comparatively simple, and 60 for Fusion~360, MCB, CADENA-Bench and BenchCAD-2. All other settings are the same in every run. The initial output limit is 15{,}000 tokens; if generation hits it before the driver can extract a command, the driver discards the partial reply and requests a new one with up to 30{,}000 output tokens. A command may attach at most 8 renders, and the conversation keeps at most 12, from the 6 most recent image-producing commands. Command outputs stay in full for 4 commands and are then replaced by their summaries. For the reported runs the model was served with tensor parallelism over four GPUs and a 250k-token context. The context actually used is set by the driver on every call (Sec.~\ref{sec:method:loop}). Mesh repair merges duplicate vertices, removes degenerate and duplicate faces, makes normals consistent and closes T-junction cracks.

\section{Experiments}
\label{sec:experiments}

\subsection{Setup}
\label{sec:exp:setup}

\paragraph{Benchmarks.}
We evaluate on six benchmarks. The first five are DeepCAD~\citep{deepcad} (300 parts), Fusion~360~\citep{fusion360} (300), MCB~\citep{mcb} (300), BenchCAD~\citep{benchcad} (1000) and CADENA-Bench~\citep{cadena} (60). The sixth is the 23 parts of task~3 of BenchCAD-2~\citep{benchcad2blog}, which is scored with BenchCAD's own part metric (Sec.~\ref{sec:exp:benchcad2}). The DeepCAD, Fusion~360 and MCB subsets are stratified by face count: each test split is divided into face-count terciles, and 100 parts are drawn at random from each tercile. The BenchCAD subset is a uniform random sample. For CADENA-Bench, 10 parts are drawn at random from each of its six part families, among parts between the 5th and 95th percentile of face count. All methods in Table~\ref{tab:main} are evaluated on the same parts. On CADENA-Bench, 4 of the 60 reference meshes were repaired before scoring. Results without them are in the supplementary material (Sec.~B).

\paragraph{Metrics.}
Each prediction is compared with the reference mesh in the reference frame, after both are normalised by the reference's bounding box (centred, longest side scaled to 1). We report volumetric IoU, which falls back to a voxel estimate when a mesh is not watertight, and the Generalized Match Score (GMS)~\citep{cadena}, which compares surface types rather than occupied volume. We also report the symmetric Chamfer distance (CD, $\times 10^3$, median over parts) on 8192 sampled points. IoU and GMS are averaged over the parts for which a method returns geometry. The invalid rate (IR) is the share of parts without geometry. On BenchCAD-2 we use the benchmark's own part score\footnote{\texttt{part\_v1} in the BenchCAD-2 evaluation code.}: $0.8\times$ an IoU term, a chance-corrected voxel IoU on a $128^3$ grid with each shape normalised to its own bounding box, plus $0.2\times$ a topology term that compares the Betti numbers (bodies, through-holes and cavities) of prediction and reference.

\paragraph{Baselines.}
We compare with CADENA~\citep{cadena}, CADFather~\citep{cadfather}, StepCAD~\citep{stepcad} and CADFit~\citep{cadfit}. The prediction meshes of CADENA, CADFather and StepCAD are rescored with our evaluation code, so that all methods are measured identically. CADFit releases no prediction meshes, so its per-part metrics are imported; it was run under the same fixed per-part compute budget as in the CADENA evaluation~\citep{cadena}, and parts whose fit did not converge within it count as invalid, which makes its means over parts with geometry optimistic. We ran StepCAD ourselves with its released inference code in its 10-minute mode (600\,s per shape); it was not evaluated on BenchCAD.

\paragraph{Protocol.}
All runs use the configuration of Sec.~\ref{sec:method:impl}, one attempt per part, and the program the agent itself returns.

\subsection{Main results}
\label{sec:exp:main}

\begin{table*}[t]
\centering
\footnotesize
\setlength{\tabcolsep}{1.7pt}
\ifdefined\tableonehook\tableonehook\fi
\begin{tabular}{l cccc@{\hspace{4pt}}cccc@{\hspace{4pt}}cccc@{\hspace{4pt}}cccc@{\hspace{4pt}}cccc}
\toprule
& \multicolumn{4}{c}{DeepCAD (300)} & \multicolumn{4}{c}{Fusion~360 (300)} & \multicolumn{4}{c}{MCB (300)} & \multicolumn{4}{c}{BenchCAD (1000)} & \multicolumn{4}{c}{CADENA-Bench (60)} \\
\cmidrule(lr){2-5}\cmidrule(lr){6-9}\cmidrule(lr){10-13}\cmidrule(lr){14-17}\cmidrule(lr){18-21}
Method & IoU & GMS & CD & IR & IoU & GMS & CD & IR & IoU & GMS & CD & IR & IoU & GMS & CD & IR & IoU & GMS & CD & IR \\
\midrule
CADENA & 0.949 & 0.958 & 0.135 & 0.67 & 0.905 & 0.896 & 0.139 & 2.00 & 0.827 & 0.660 & 0.302 & 0.33 & 0.927 & 0.879 & 0.155 & 0.70 & 0.895 & 0.762 & 0.255 & \textbf{0.00} \\
CADFather & 0.980 & 0.979 & 0.130 & \textbf{0.00} & 0.960 & 0.943 & 0.120 & \textbf{0.00} & 0.862 & 0.705 & 0.260 & \textbf{0.00} & 0.967 & 0.910 & 0.145 & \textbf{0.00} & 0.918 & 0.800 & 0.240 & \textbf{0.00} \\
StepCAD & 0.884 & 0.844 & 0.228 & \textbf{0.00} & 0.775 & 0.732 & 0.281 & \textbf{0.00} & 0.674 & 0.495 & 1.001 & \textbf{0.00} & -- & -- & -- & -- & 0.754 & 0.582 & 0.800 & \textbf{0.00} \\
CADFit$^\dagger$ & 0.863 & 0.826 & 0.175 & 23.00 & 0.786 & 0.793 & 0.244 & 18.67 & 0.679 & 0.545 & 0.614 & 24.33 & -- & -- & -- & -- & 0.696 & 0.592 & 0.553 & 3.33 \\
CADMAGEDDON & \textbf{$>$0.999} & \textbf{0.991} & \textbf{0.128} & \textbf{0.00} & \textbf{0.993} & \textbf{0.973} & \textbf{0.112} & 0.33 & \textbf{0.933} & \textbf{0.818} & \textbf{0.196} & 0.67 & \textbf{0.999} & \textbf{0.975} & \textbf{0.137} & 0.30 & \textbf{0.981} & \textbf{0.909} & \textbf{0.172} & \textbf{0.00} \\
\bottomrule
\end{tabular}
\caption{Reconstruction quality on five benchmarks (number of parts in parentheses). Higher IoU and GMS and lower CD and IR are better. IoU and GMS are averaged over parts with geometry; on BenchCAD, IoU is our evaluation code's, not BenchCAD's $64^3$ voxel IoU (Table~\ref{tab:leaderboard}). CD is the median Chamfer distance ($\times 10^3$). IR is the share of parts without geometry (\%). All methods except CADFit are scored by our evaluation code from their prediction meshes. $^\dagger$CADFit releases no meshes, so its imported per-part metrics are used.}
\label{tab:main}
\end{table*}

Table~\ref{tab:main} compares CADMAGEDDON with the baselines. It reaches the highest IoU and GMS and the lowest CD on all five benchmarks. The margin over the strongest baseline, CADFather, is smallest on DeepCAD and grows on the mechanical-part sets. IoU rises from 0.862 to 0.933 on MCB and from 0.918 to 0.981 on CADENA-Bench. The largest gains are in GMS, the metric that compares surface types: from 0.705 to 0.818 on MCB and from 0.800 to 0.909 on CADENA-Bench. This is the surface-structure gap identified in Sec.~\ref{sec:intro}. CD on MCB falls from 0.260 to 0.196. CADMAGEDDON leaves at most 0.67\% of parts without geometry, while CADFit fails on 18.7--24.3\% of DeepCAD, Fusion~360 and MCB.

\subsection{Comparison with frontier models on BenchCAD}
\label{sec:exp:leaderboard}

\begin{table}[!t]
\centering
\small
\setlength{\tabcolsep}{2.5pt}
\begin{tabular}{llrrr}
\toprule
Method & Input & Parts & IoU$\uparrow$ & IR\%$\downarrow$ \\
\midrule
Claude Sonnet 5.5 (max)$^\ast$ & img. & 1000 & 0.747 & -- \\
Claude Opus 5.5 (max)$^\ast$ & img. & 1000 & 0.730 & -- \\
GPT-5.6 Sol (max)$^\ast$ & img. & n/s & 0.706 & -- \\
Gemini 3.1 Pro (think.) & img. & n/s & 0.355 & 18.5 \\
Claude Opus 4.7 (think.) & img. & n/s & 0.279 & 3.5 \\
\midrule
Claude Sonnet 5.5 (max)$^\ast$ & img.+tools & 1000 & 0.963 & -- \\
Claude Opus 5.5 (max)$^\ast$ & img.+tools & 1000 & 0.962 & -- \\
GPT-6 Astra$^\ast$ & img.+tools & 1000 & 0.959 & -- \\
GPT-5.6 Sol (max)$^\ast$ & img.+tools & n/s & 0.834 & -- \\
\midrule
CADENA-RL~\citep{cadena} & mesh & 17\,900 & 0.910 & 0.9 \\
CADFather~\citep{cadfather} & mesh & 1000 & 0.968 & \textbf{0.0} \\
CADMAGEDDON & mesh & 1000 & \textbf{0.991} & 0.3 \\
\bottomrule
\end{tabular}
\caption{BenchCAD leaderboard. The frontier rows are taken from the BenchCAD leaderboard~\citep{benchcadboard}; $^\ast$marks results the vendor reported itself, without IR. Input: images (img.) or the mesh; with tools, a model can also run Python to render, measure and revise before it submits. Parts is the number of the 17\,900 BenchCAD parts a row is evaluated on (n/s: not stated). CADENA-RL and CADFather are reported in their papers. CADMAGEDDON is scored with BenchCAD's own $64^3$ voxel IoU: its programs are built to STEP exactly as BenchCAD scores the code of LLM baselines, and a part without a prediction counts as 0.}
\label{tab:leaderboard}
\end{table}

Table~\ref{tab:leaderboard} places CADMAGEDDON on the BenchCAD leaderboard. Frontier models that receive images and a Python sandbox report up to 0.963 IoU on 1000-part subsets. Without the sandbox they reach at most 0.747. CADMAGEDDON, driven by the mid-sized Qwen3.8-27B and scored with the same $64^3$ voxel IoU, reaches 0.991 on a 1000-part subset with 0.3\% invalid predictions. Scored with the same $64^3$ voxel IoU from the delivered meshes on the same 1000 parts, it reaches 0.995, against 0.968 for CADFather and 0.930 for CADENA.

\begin{figure*}[t]
\centering
\includegraphics[width=0.76\textwidth]{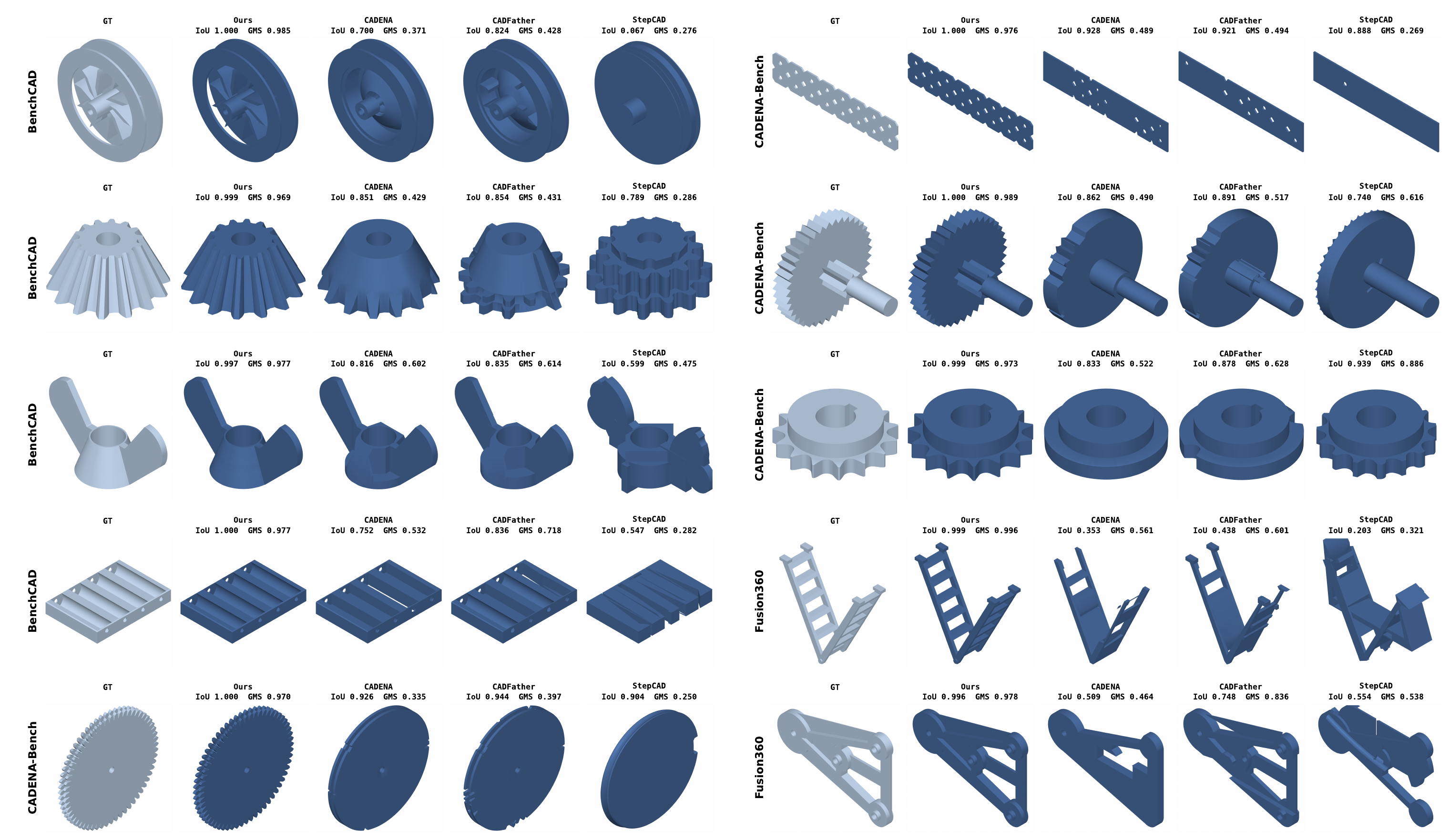}
\caption{Reconstructions of BenchCAD, CADENA-Bench and Fusion~360 parts by CADMAGEDDON (Ours), CADENA, CADFather and StepCAD, with the reference (GT) on the left of each row. Numbers are IoU and GMS. The parts are those on which CADMAGEDDON's margin over the best baseline is largest. StepCAD was run on the four BenchCAD parts for this figure only.}
\label{fig:qualitative}
\end{figure*}

\subsection{BenchCAD-2 and FreeCAD}
\label{sec:exp:benchcad2}

\begin{table}[!t]
\centering
\footnotesize
\setlength{\tabcolsep}{2.5pt}
\begin{tabular}{l rrrr rr}
\toprule
Method & Score & IoU & Topo. & sF1 & $\geq$0.9 & \$/part \\
\midrule
\multicolumn{7}{l}{\emph{Leaderboard, four rendered views, 28 cases$^\P$}} \\
\quad GPT-6 Astra & 0.58 & -- & -- & -- & -- & -- \\
\quad Claude Opus 5.5 & 0.59 & -- & -- & -- & -- & -- \\
\quad Claude Sonnet 5.5 & 0.59 & -- & -- & -- & -- & -- \\
\quad Claude Haiku 5.5 & 0.32 & -- & -- & -- & -- & -- \\
\quad Claude Fable 5.1 & 0.31 & -- & -- & -- & -- & -- \\
\midrule
\multicolumn{7}{l}{\emph{Mesh input, 23 parts}} \\
Qwen3.8, Qwen Code$^\circ$ & 0.790 & 0.804 & 0.856 & 0.823 & 17 & 2.71 \\
\midrule
\multicolumn{7}{l}{\emph{CADMAGEDDON}} \\
\quad CadQuery & \textbf{0.933} & \textbf{0.928} & \textbf{0.957} & \textbf{0.946} & \textbf{20} & \textbf{1.05} \\
\quad FreeCAD  & 0.925 & 0.927 & 0.920 & 0.945 & \textbf{20} & 1.27 \\
\bottomrule
\end{tabular}
\caption{BenchCAD-2, task~3 (23 parts), scored with BenchCAD's part score (Sec.~\ref{sec:exp:setup}); no prediction counts as 0. IoU: IoU term; Topo.: topology term; sF1: surface F1; $\geq$0.9: parts with score $\geq 0.9$. \$/part: API cost at OpenRouter prices without cache discount (Qwen3.8-27B \$0.035/\$1.5 per 1M input/output tokens, its lowest listed provider), counting all attempts not cut short by an outage. $^\P$BenchCAD~2.0 leaderboard, reverse-engineering part task~\citep{benchcad2blog}: four rendered views, max effort, harness v2.0, on the 28-case private set; our 23 parts are the public set. $^\circ$Qwen Code harness~\citep{qwencode}, same model as ours, best of two attempts; 2 parts without prediction.}
\label{tab:benchcad2}
\end{table}

On the BenchCAD~2.0 leaderboard, frontier models reconstruct this task's parts from four rendered views, and none exceeds a part score of 0.59~\citep{benchcad2blog}. With the mesh as input, CADMAGEDDON reaches 0.933 with CadQuery at \$1.05 per part (Table~\ref{tab:benchcad2}). The same Qwen3.8-27B in a general-purpose agent harness, Qwen Code, shows what the instruments contribute: as the best of two attempts it reaches a part score of only 0.790, returns no prediction on 2 of the 23 parts, and needs about 18M input tokens per attempt.

An agent that uses the same visualisation and detection instruments but emits FreeCAD PartDesign programs through its own DSL is on par with the CadQuery version under the same model, call limit and selection: 0.925 vs.\ 0.933, and both reach a part score of at least 0.9 on 20 of the 23 parts. The reconstructions can therefore also be delivered as native FreeCAD models.

\subsection{Qualitative comparison}
\label{sec:exp:qualitative}

Fig.~\ref{fig:qualitative} shows the parts with the largest margins. On gears, sprockets, the bevel gear and the impeller, the baselines reproduce the overall disc, cone or hub but merge or omit the teeth and blades, while CADMAGEDDON recovers them. On the perforated strip and the ladder-like Fusion~360 part, the baselines fill in or simplify the openings. CADMAGEDDON reconstructs them as separate features, which is what the GMS gap in Table~\ref{tab:main} measures.

Analyses of the call budget and of tool use can be found in the supplementary material (Sec.~A).

\section{Limitations}
\label{sec:limitations}

CADMAGEDDON assumes a clean mesh of a single part: preprocessing repairs cracks and duplicate faces, not scan noise or missing regions, and all our benchmarks are derived from CAD models. Assemblies and real scans are left to future work.

The instruments bound what the agent can express: a feature that no detector names or no operation produces can only be approximated. On BenchCAD-2, for example, we observe parts whose 0.7\,mm thin walls are left out of the reconstruction, and a part with 51 through-holes of which at most two are built.

The emitted CadQuery can be edited, but its dimensions are literal values rather than named parameters, and fillets and chamfers are explicit solids rather than edge operations; recovering design intent is beyond the current system.

Our comparison with frontier models relies on leaderboard entries that differ in input modality and evaluated subsets. A systematic comparison that separates the contributions of the agent harness, the specific frontier model and the input modality is left to future work; its cost grows quickly with the number of models, harnesses and benchmarks. Finally, the agent trades model size for interaction: a reconstruction takes up to 30 or 60 model calls with long contexts, so it is slower than single-pass learned models.

\section{Conclusion}
\label{sec:conclusion}

CADMAGEDDON reconstructs editable CAD programs from meshes with one mid-sized open-weight model run on premises. The model chooses each feature's structure among named geometric entities, and fitting supplies its values. With Qwen3.8-27B it leads on all five benchmarks of Table~\ref{tab:main}, most clearly in the surface structure of mechanical parts, and writes equally good FreeCAD programs. On BenchCAD-2, the same mid-sized model in a general-purpose agent harness falls well behind. The instruments, not model scale, carry much of the accuracy.

\bibliographystyle{ieeenat_fullname}
\bibliography{refs}

\clearpage
\appendix
\begin{center}{\Large\bfseries Supplementary Material}\end{center}
\section{Analysis}
\label{sec:supp:analysis}
\label{sec:exp:ablations}

\begin{figure*}[t]
\centering
\includegraphics[width=0.4\textwidth]{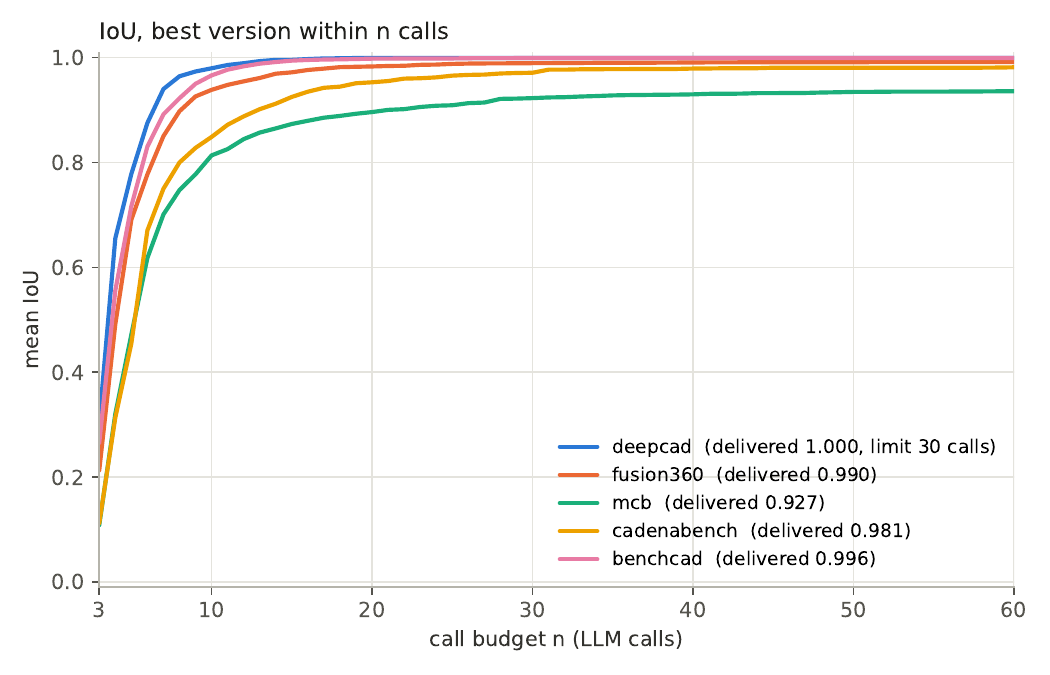}\qquad
\includegraphics[width=0.4\textwidth]{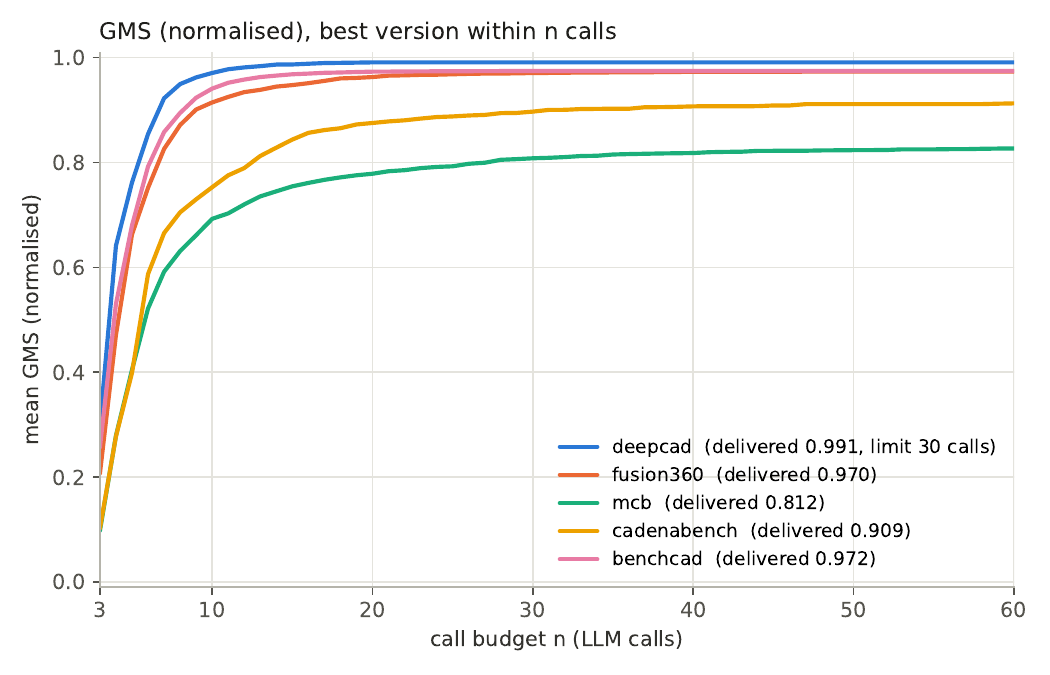}
\caption{IoU (left) and GMS (right) of the best program version produced within the first $n$ model calls, per benchmark. Parts with no version yet count as 0. The best version is chosen by the plotted metric, i.e.\ with the reference, so each curve is an upper bound on what a run stopped at $n$ calls could return. The legend gives the mean of the program the agent itself delivered at its full budget (30 calls on DeepCAD and BenchCAD, 60 otherwise); like the curves, it counts parts without a program as 0, so it can differ slightly from Table~1 of the main paper.}
\label{fig:budget}
\end{figure*}

\subsection{Call budget}
Fig.~\ref{fig:budget} shows how quality grows with the number of model calls. On DeepCAD and BenchCAD the best program within 10 calls already reaches an IoU of 0.980 and 0.966, and both curves are flat after 20 calls, which is why a budget of 30 suffices there. The mechanical-part sets need longer runs. On MCB the IoU is 0.814 after 10 calls, 0.896 after 20 and 0.923 after 30. On CADENA-Bench it is 0.849, 0.953 and 0.971. GMS rises more slowly than IoU on both sets and is still increasing at 60 calls. This is consistent with surface detail, which GMS measures, being added by later features such as holes, blends and patterns.

\begin{figure*}[t]
\centering
\begin{minipage}[c]{0.49\textwidth}
\centering
\includegraphics[width=0.92\linewidth]{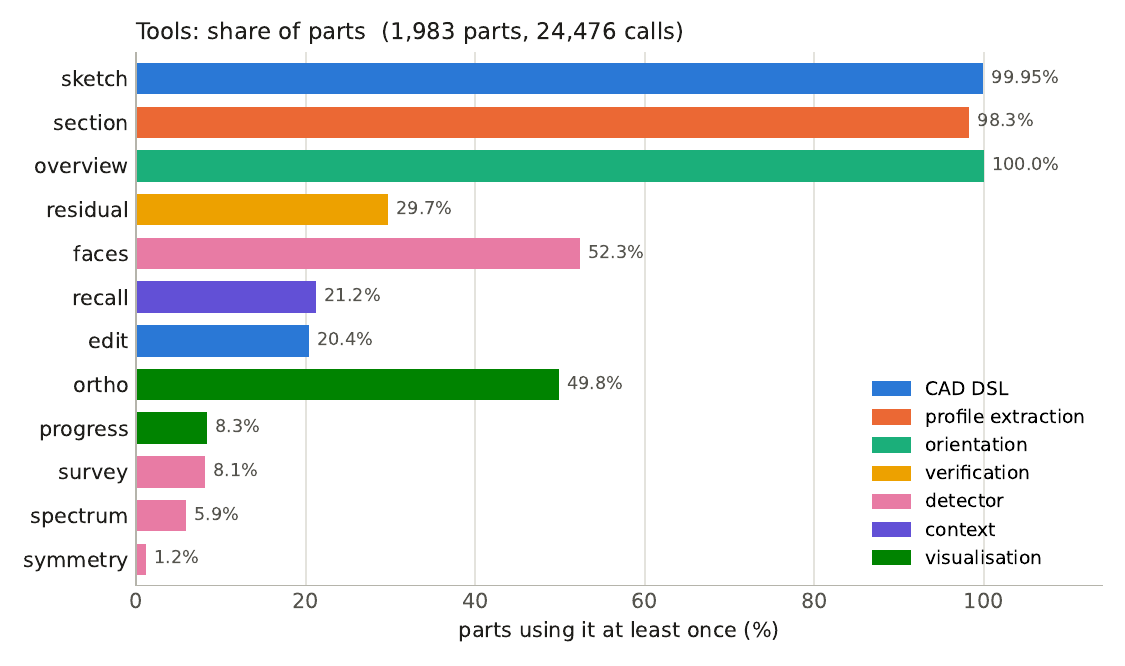}\\[2pt]
\includegraphics[width=0.92\linewidth]{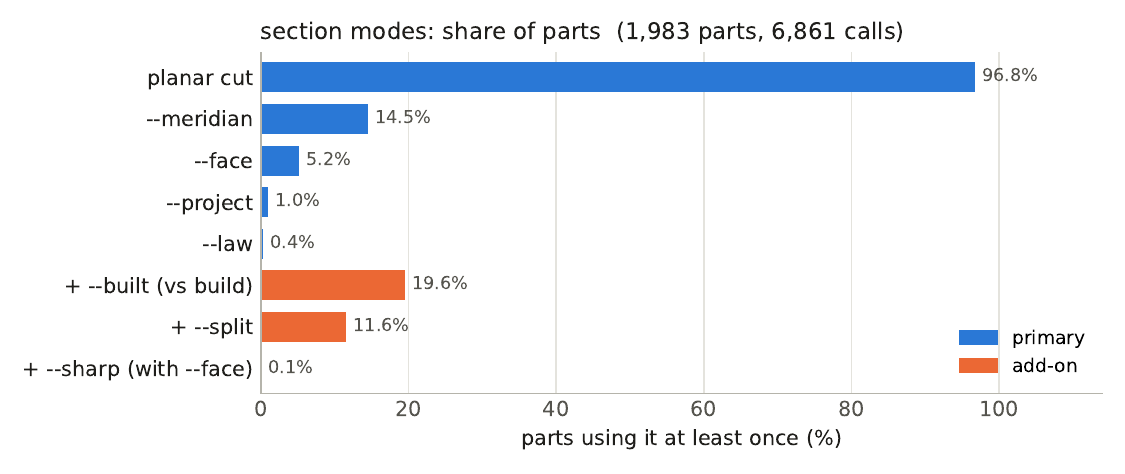}
\end{minipage}\hfill
\begin{minipage}[c]{0.49\textwidth}
\centering
\includegraphics[width=0.92\linewidth]{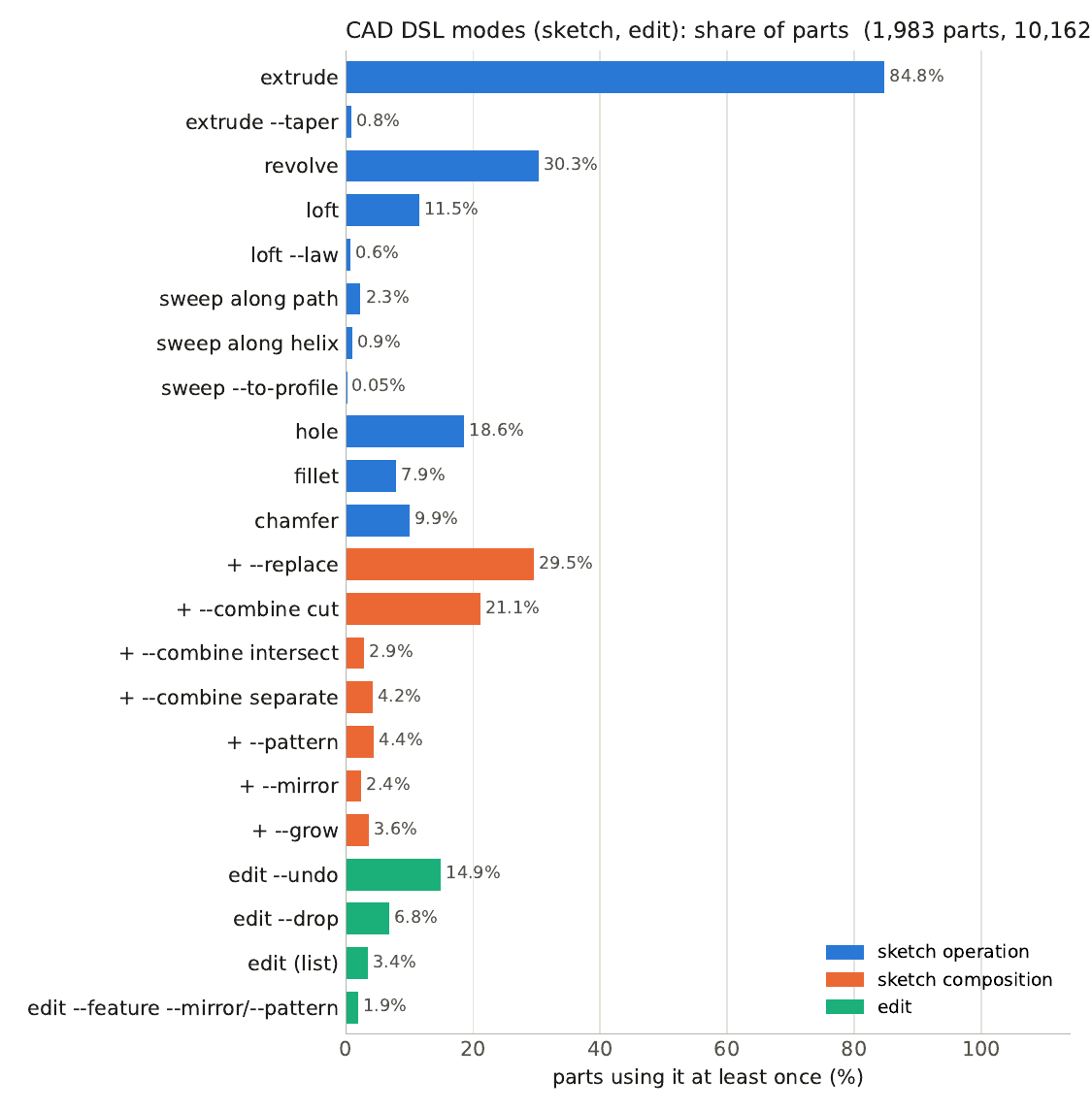}
\end{minipage}
\caption{Tool use. Share of parts on which each instrument (top left), each mode of \texttt{section} (bottom left) and each construction or editing mode (right) is used at least once. The statistics cover 1983 parts and 24\,476 calls over the delivered attempts of all six runs. On BenchCAD-2 the attempt with the best reference score is used, as in Table~3 of the main paper. Top left: colours give the instrument's role. Bottom left and right: primary modes and operations in blue, added options in orange, edits of existing features in green.}
\label{fig:tools}
\end{figure*}

\subsection{Tool use}
Fig.~\ref{fig:tools} shows which instruments the agent relies on. Nearly every reconstruction follows the core loop: \texttt{overview} (100\% of parts), \texttt{section} (98.3\%) and \texttt{sketch} (99.95\%). The other instruments are used selectively. Analytic faces and hole stacks (\texttt{faces}) are listed on 52.3\% of parts, elevations with named levels (\texttt{ortho}) on 49.8\%, and the \texttt{residual} instrument on 29.7\%. \texttt{survey}, \texttt{spectrum} and \texttt{symmetry} are each used on fewer than 10\%. Among construction operations, extrusion dominates (84.8\% of parts), followed by revolution (30.3\%), holes (18.6\%) and lofts (11.5\%). Sweeps along paths and helices are rare (2.3\% and 0.9\%). Correction is common. On 29.5\% of parts the agent replaces a feature it built, on 14.9\% it undoes one, and on 6.8\% it drops one. The reconstruction is thus not a single forward pass: the agent both constructs and corrects.

\section{CADENA-Bench without repaired references}
\label{sec:supp:repaired}

Four of the 60 CADENA-Bench reference meshes are not valid solids and were repaired before scoring (Sec.~4.1 of the main paper). Table~\ref{tab:repaired} repeats the CADENA-Bench comparison of Table~1 of the main paper on the remaining 56 parts. The ranking is unchanged.

\begin{table}[h]
\centering
\small
\setlength{\tabcolsep}{4pt}
\begin{tabular}{l rrrr}
\toprule
Method & IR (\%) $\downarrow$ & IoU $\uparrow$ & GMS $\uparrow$ & CD $\downarrow$ \\
\midrule
CADENA~\citep{cadena} & \textbf{0.00} & 0.890 & 0.759 & 0.255 \\
CADFather~\citep{cadfather} & \textbf{0.00} & 0.919 & 0.806 & 0.231 \\
StepCAD~\citep{stepcad} & \textbf{0.00} & 0.769 & 0.593 & 0.800 \\
CADFit$^\dagger$~\citep{cadfit} & 1.79 & 0.696 & 0.622 & 0.424 \\
CADMAGEDDON & \textbf{0.00} & \textbf{0.981} & \textbf{0.912} & \textbf{0.168} \\
\bottomrule
\end{tabular}
\caption{CADENA-Bench without the 4 repaired references (56 parts); metrics as in Table~1 of the main paper. $^\dagger$Imported per-part metrics.}
\label{tab:repaired}
\end{table}

\end{document}